\documentclass[fleqn,10pt]{wlscirep}
\usepackage[utf8]{inputenc}
\usepackage[T1]{fontenc}
\usepackage{lineno}

\usepackage{siunitx}  
\usepackage{caption}
\usepackage{subcaption}
\usepackage{amsmath}
\usepackage{colortbl}
\usepackage{makecell}

\usepackage{pifont}
\newcommand*\colourcheck[1]{%
  \expandafter\newcommand\csname #1check\endcsname{\textcolor{#1}{\ding{52}}}%
}
\newcommand*\colourxmark[1]{%
  \expandafter\newcommand\csname #1xmark\endcsname{\textcolor{#1}{\ding{55}}}%
}
\colourcheck{green}
\colourxmark{red}

\usepackage{booktabs}
\usepackage{dcolumn}
\newcolumntype{d}[1]{D..{#1}}

\usepackage{caption}
\title{Language models trained on media diets can predict public opinion}

\author[$\dagger$]{Eric Chu \footnote{Work done at MIT, now at Google.}}
\author[1]{Jacob Andreas}
\author[2]{Stephen Ansolabehere}
\author[1]{Deb Roy}
\affil[1]{Massachusetts Institute of Technology}
\affil[2]{Harvard University}

\affil[$\dagger$]{corresponding author(s): Eric Chu (eric.chu.56@gmail.com)}

\begin{abstract}
Public opinion reflects and shapes societal behavior, but the traditional survey-based tools to measure it are limited.
We introduce a novel approach to probe \textit{media diet models} -- language models adapted to online news, TV broadcast, or radio show content -- that can emulate the opinions of subpopulations that have consumed a set of media. To validate this method, we use as ground truth the opinions expressed in U.S. nationally representative surveys on COVID-19 and consumer confidence. Our studies indicate that this approach is 
(1) predictive of human judgements found in survey response distributions and robust to phrasing and channels of media exposure, 
(2) more accurate at modeling people who follow media more closely, and 
(3) aligned with literature on which types of opinions are affected by media consumption.
Probing language models provides a powerful new method for investigating media effects, 
has practical applications in supplementing polls and forecasting public opinion, 
and suggests a need for further study of the surprising fidelity with which neural language models can predict human responses.
\end{abstract}
\begin{document}

\flushbottom
\maketitle

\thispagestyle{empty}

\section*{Introduction}

Mass media controls the diffusion of information in our society,  \cite{doi:10.1080/10584600802426965} frames key political and economic issues, \cite{lakoff2014all} shapes assessments of political figures, \cite{10.2307/2938828} and impacts public health. \cite{berry2007sars}
The effects of media on our society ripple beyond the individual, giving rise to new social movements \cite{greenberg2004framing} and determining national agendas. \cite{king2017news} 
Social scientists have long studied media with an eye to the normative implications for the health of democratic governance and are increasingly concerned about whether misinformation, fake news, \cite{lazer2018science} and echo chambers \cite{flaxman2016filter} are wearing away at the public ethos. \cite{allen2020evaluating}
A rich literature using experiments and surveys has examined the impact of media exposure on people’s attitudes and behaviors, yet there is still a gap between the actual content of the news and the questions asked in surveys.
Here we ask whether recent advances in deep learning and language modeling offer a new approach to predict public opinion.

Public opinion – sometimes referred to as a `thermostat' of public will \cite{10.2307/2111666, stimson1995dynamic} – is commonly measured through surveys by governments, companies, NGOs, and political parties and candidates for office.\cite{berinsky2017measuring} The understanding gained through surveys is critical inputs to decision-making around economic strategy and public health. Shifts in polling results have also been tied to shifts in public policy \cite{10.2307/1956018} and can reflect impending societal changes. However, polling faces several problems. In 1997, Pew was able to receive responses to 36\% of potential respondents, but that response rate had fallen to only 9\% by 2012. Shifting to Internet-based polling has helped alleviate this issue partially, but it can also be harder to sample representative populations online. \cite{berinsky2017measuring} In both online and offline settings, polls can be expensive, with a typical national survey of adult Americans ranging from \$10 to \$1,000 per interview. \cite{ansolabehere2018taking} 
Our results suggest the possibility of using media diet models to supplement public opinion polls by emulating survey respondents, and to forecast shifts in public opinion.

Despite theory and expectation of large media effects, \cite{lippmann2017public, valkenburg2013comm} media effect studies have found small to moderate effect sizes. \cite{valkenburg2013comm, valkenburg2016media} Attenuated effect sizes have been attributed to (a) \textit{media content} not being incorporated, and (b) \textit{media exposure} being only loosely measured.
For example, a typical media impact study might assess media effects by measuring the correlation between answers to ``how many hours of online news do you consume per week'' and ``how concerned are you about domestic terrorism, on a five-point scale''. Here, the news sources are missing, and the media content is left out of the analysis completely.
More recent work has attempted to address the content modeling issue by counting keywords, computing the average sentiment, or extracting topics from media coverage \cite{o2010tweets, grimmer2013representational}; however, these are only coarse summaries of media messaging.
Our approach for public opinion prediction is designed with these problems in mind.
Neural language models can better capture the semantics of \textit{media content}, and adapting them to specific sources can be tuned for \textit{media exposure}; 
this method also allows one to trace predictions for a survey question back to the media content that may have influenced it.

Developments in AI have spurred progress in domains as varied as speech recognition, genomics, and visual object recognition. \cite{lecun2015deep} In natural language processing (NLP), many recent successes in have been propelled by large, neural language models trained on massive corpora.
Several of these language models are trained through a ``masked language modeling'' (MLM) objective, in which the model must learn to ``fill-in-the-blank''. \cite{devlin-etal-2019-bert}
For example, given the sentence with missing words: ``The Pyramid of Nyuserre [BLANK] a [BLANK] complex [BLANK] in the 25th century BC for the Egyptian pharaoh Nyuserre Ini of the [BLANK] Dynasty'', the model must learn high-dimensional vector representations that can be used to predict `is', `pyramid', `built', and `Fifth'. This approach allows the model to learn representations of linguistic structure (`is' often follows a noun and precedes `a'), contextual co-occurrences (`pyramid' is likely given the previous mention of pyramid and other Egyptian references), semantic information (pyramids are structures that are `built'), and general world knowledge (Ini was a pharaoh in the `Fifth' Dynasty).
MLM models have become foundational for much of NLP, and have achieved impressive results in question answering, machine translation, natural language inference, and many other tasks.
Research on probing these models has largely focused on assessing and extracting factual knowledge, \cite{petroni2019language} but our method extends this concept to predict subpopulation-specific opinions.

Our research raises several questions centered around AI’s ability to mirror and mimic beliefs derived from human language.
Recent work such as GPT3, \cite{brown2020language} PaLM, \cite{chowdhery2022palm}, ChatGPT, Claude, and Bard have mainstreamed public awareness of large language models, and answering these questions has become more urgent as models continue to improve and become more widely used.
In this work, our broad hypothesis is that language models can predict public opinion. We show that even simpler N-gram -based models are capable of this to some degree, though using a more powerful model such as BERT \cite{devlin-etal-2019-bert} greatly increases the accuracy. This trend may continue with the even more powerful large language models mentioned above, which are built using fundamentally similar technology and model architectures.

We investigate whether human survey responses can be approximated by language models trained on particular media diets. To create \textit{media diet models}, we start with a language model (e.g. the popular BERT model), and finetune it on a media diet dataset. This adaptation allows the model to absorb new information, while also updating internal knowledge representations already present in BERT.
We then probe these models with questions, and examine whether they match survey response distributions of subpopulations with specific media diets.
This work is in the line of new survey methodology  \cite{10.2307/2998167} and modeling. \cite{bartels1993messages}
Similar to observational studies and natural experiments, \cite{hennighausen2015exposure, 10.2307/4620110, dellavigna2007fox} this opens up the study of messaging in the wild.
The media diet models are shown to have predictive power across public health and economic settings, be robust to phrasing of questions and effective across media sources, and contain predictive signal even when controlling for the demographics of each subpopulation.
Further analyses illustrate their sensitivity to how closely people are paying attention to news, and heterogeneous effects dependent on question type.

\subsection*{Media diet modeling approach}

\begin{figure}[!ht]
\centering
\includegraphics[width=1.0\linewidth]{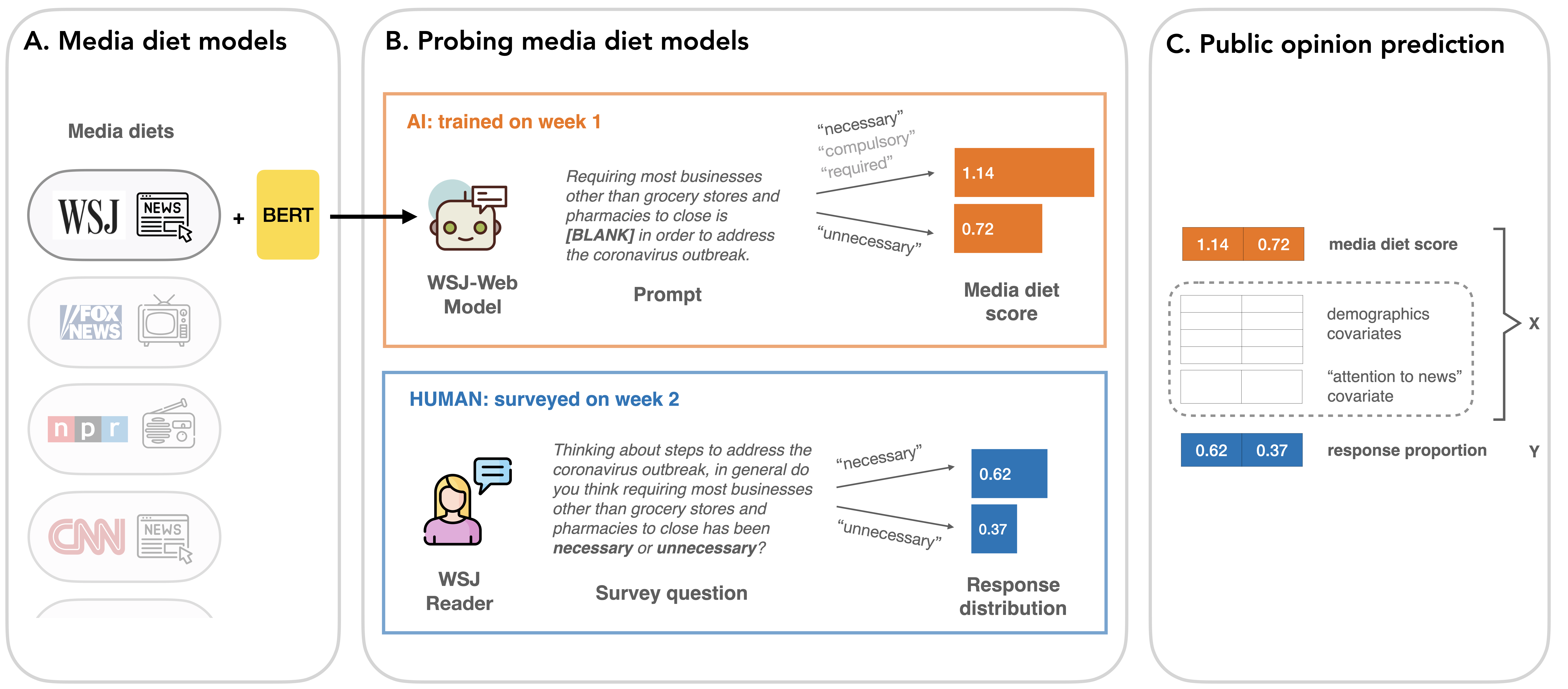}
\small
\caption{
Overview of media diet modeling approach.
(A) One media diet dataset is textual data from one or more sources, from a particular medium (online, TV, radio), over a given time period.
A language model such as BERT is adapted to a dataset to create one \textit{media diet model}.
(B) 
The top row illustrates the calculation of a \textit{media diet score} for a given fill-in-the-blank prompt and target word. In this example, the target words are `necessary' and `unnecessary'. In our synonym-grouping method, we aggregate probabilities for synonyms of the target word -- `compulsory', `required', etc. -- in this calcuation. 
In our studies, prompts are derived from real survey questions asked in national polls. 
In the bottom row, the original survey question and its two possible choices -- `necessary' and `unnecessary' -- is shown. The response distribution shows that out of the people who listed WSJ as their primary news source, 62\% answered `necessary', versus 37\% for `unnecessary'.
Notably, the models are adapted to news one week/month before the time the survey was conducted.
(C) Our hypothesis is that the target word probabilities, which are updated after finetuning BERT, reflect media effects. These in turn are predictive of the response distributions found in surveys.
The media diet scores are used to predict the response proportions, combining data over multiple media diets and surveys.
In additional analyses, we include demographic stats and information about how closely respondents were paying attention to news.
}
\normalsize
\label{fig:overall_fig}
\end{figure}

The main idea behind our approach is to build a computational model that takes as input a description of an subpopulation's media diet, and a survey question, and produces as output a prediction of how the subpopulation will respond to the survey question. 
If this model predicts real human survey judgments well, there is potential to use it as an \textit{in silico} model of public opinion. 
This could help answer public opinion questions (``how do people feel about the pandemic''), as well as scientific questions around media effects (``how does media diet affect perceptions of the pandemic''). The overall approach is shown in Figure \ref{fig:overall_fig}.

Building a media diet model involves three steps. In step one, we create or use a base language model that can predict missing words in text. We use pretrained models in our work, with BERT as our main model.
In step two, we adapt the language model by \textit{fine-tuning} it on a specific media diet dataset, which contains media content from one or a mixture of news sources from a given time period. We use online news articles, TV transcripts, and radio show transcripts.

In step three, we query the media diet model and score answers to survey questions. Throughout the rest of this paper, we demonstrate that these scores are correlated with human judgments. Here we mean that there is correlation between (i) a probability-based score that the model assigns to a given answer, and (ii) the fraction of survey participants that choose a particular answer.
To do public opinion \textit{prediction}, we fit regression models that use (i) to predict (ii). The survey data comes from national polls conducted about COVID-19 and consumer confidence. Finally, we use a nearest neighbor approach to trace predictions for a given survey question back to the original media diet datasets.

More concretely, our probing method produces a score $s$ for a language model $LM$, a fill-in-the-blank prompt, and target word $w$. We write this as $s = LM(prompt, w)$.
For the BERT baseline (no adaption to media content), this score is simply the probability of $w$ in the blank position. For instance, for the prompt ``Requiring most businesses other than grocery stores and pharmacies to close is [BLANK] in order to address the coronavirus outbreak'', the probability of `necessary' is $s=0.188$.
For a \textit{media diet model}, we divide its score by the baseline, non-adapted model score. This is: $s = \frac{MediaDietBERT(prompt, w)}{BERT(prompt, w)}$. This normalization means the score reflects new information contained in the media diet dataset, but relative to existing knowledge and information in the base model.
Finally, we introduce a \textit{synonym-grouping} scoring method, in which probabilities are summed over synonyms over the target word. This helps address the ``surface form competition'' issue, \cite{holtzman2021surface} and allows the probing to be less sensitive to the exact question phrasing and target word choice. More details are in the Methods section.

\section*{Results} \label{sec:results}

Our first study centers around the COVID-19 pandemic. Polarized news coverage and high stakes political outcomes created the need to understand how media effects shaped the course of the pandemic. Media effects on public health can be varied, ranging from the promotion of healthy practices and fostering of public debate, to sensationalizing health issues and causing unfounded fear in the general public. \cite{berry2007sars} Studies on previous epidemics, such as the SARS crisis of 2003, found news media to have disproportionately focused on risk rather than prevention \cite{berry2007sars} and to affect intentions to comply with messaging from health organizations like the Centers for Disease Control and Prevention. Since the start of the COVID-19 outbreak, multiple studies have tied high-stakes behavior and outcomes to media coverage. For example, Bursztyn et. al found that viewers of February 2020 TV broadcasts that downplayed the severity of the pandemic were 39\% more likely to contract and die from the coronavirus. \cite{NBERw27417} Others have found polarization in media to result in a decreased adherence to social distancing measures. \cite{NBERw26946, NBERw27237}

Using survey data of a nationally representative sample of U.S. adults provided by Pew Research Center, \cite{pew_pathways} we investigated the validity of our media diet models as a proxy for media diet consumption. American News Pathways is a project that conducted surveys on panels of over twelve thousand U.S. adults, with several surveys centering around respondents’ beliefs and knowledge around COVID-19. Critically, four surveys conducted across March, April, and June 2020 contain media diet information for respondents, such as which major outlet (if any) they consider as their primary source of news. We created weekly media diet models for four major outlets found in the Pew data that have online news articles and span a range of political bias -- CNN, Fox News, New York Times (NYT), and National Public Radio (NPR). Each media diet model is trained by adapting BERT to a week of coronavirus-related news articles from a week before the survey. Except where noted in certain sub-analyses, the results are using \textit{online news} -based models used to predict survey response proportions.

Our second study examines nationally representative consumer confidence surveys provided by the University of Michigan. \cite{mich_surveys} Consumer confidence has been extensively studied in economics since the inception of these surveys in the 1950's, are summarized as a consumer sentiment index used by businesses and banks, and can influence political evaluations and national economic growth. \cite{ludvigson2004consumer} Understood to be driven by both real economic conditions and media coverage, previous research has found, for example, the amount of negative news \cite{HOLLANDERS2011367} and level of uncertainty in messaging \cite{van2016mediated} to have casual effects on consumer confidence. We obtained media diet information for respondents in the University of Michigan surveys by linking them to Pew Pathways respondents, based on demographic buckets. We examined four groups with distinct media diets and create monthly media diet models for each. Each media diet model is trained by adapting BERT to a month of economy- and finance -related online, TV, and radio news from around the time of the survey. Survey responses proportions to twenty two questions, asked repeatedly every month, are compared to the scores from our media diet models. We conducted our analysis on 24 months of survey data from 2019 to 2020. In the default setting, we use \textit{online news} models to predict survey response proportions \textit{one month after} the news articles were published.

We are interested in the validity and sensitivity of our approach, which we examine through the following questions:

\textbf{RQ1a} (Effectiveness) Do the media diet models have predictive power for survey responses?

\textbf{RQ1b} (Modeling): Are pretrained, neural language models necessary, or are simpler language models sufficient?

\textbf{RQ1c} (Modeling): Is the synonym-grouping method in our probing approach necessary?

\textbf{RQ1d} (Robustness): Are the media diet models robust to paraphrases of survey question prompts?

\textbf{RQ1e} (Robustness): Do the media diet models have predictive power across media sources?

\textbf{RQ1f} (Robustness): Do media diets have predictive power, even when controlling for demographics of each subpopulation?

\textbf{RQ2} (Media exposure) Are the media diet models sensitive to the amount of attention being used to news? Are they more accurate when people are paying more attention to media?

\textbf{RQ3} (Media effects) Is the method more effective for certain topics or types of opinions?

\subsection*{Domain 1: Attitudes Towards COVID-19}

\begin{figure}[!h]
\centering
\includegraphics[width=1.0\linewidth]{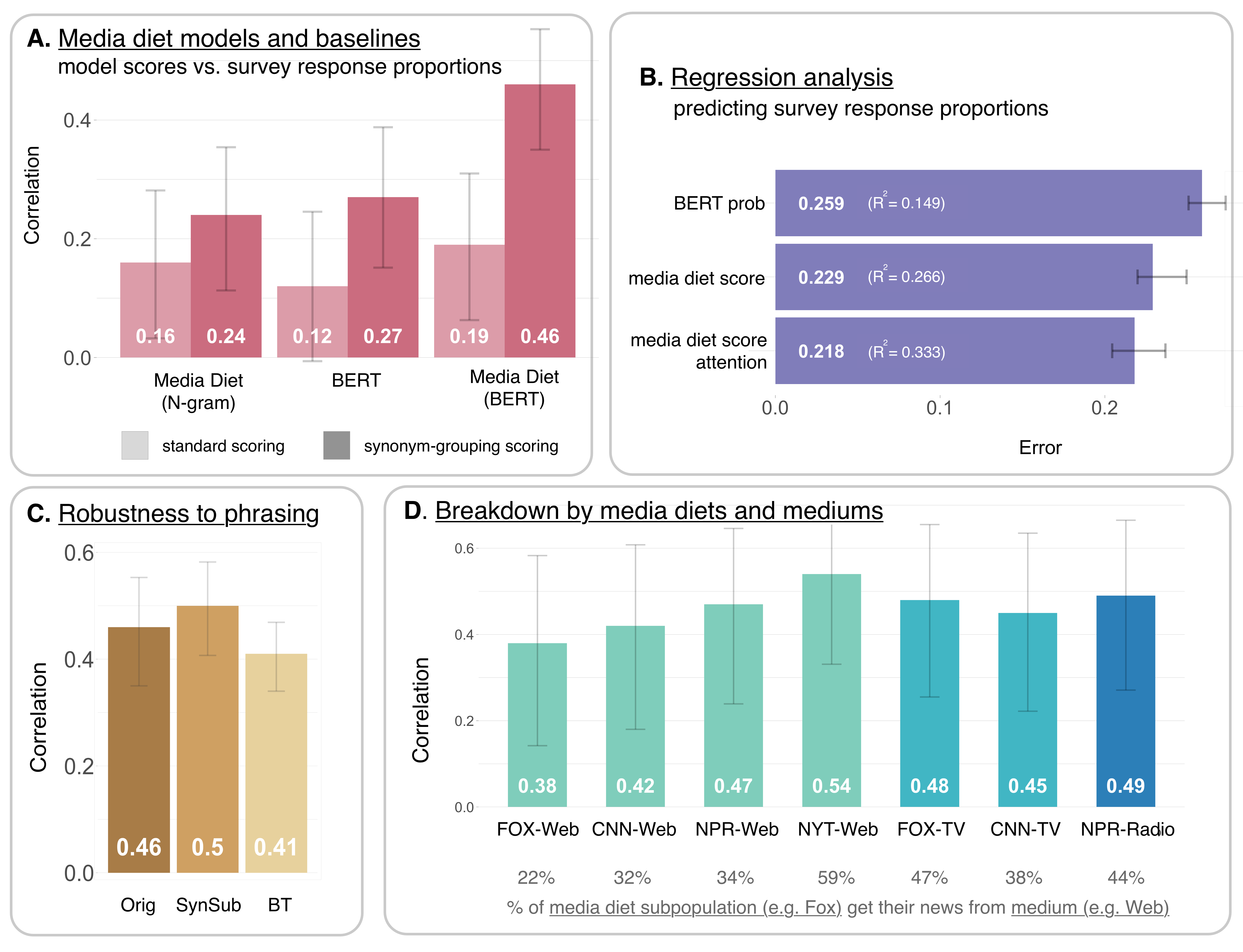}
\small
\caption{
Attitudes towards COVID-19: correlations and regressions on media diet scores and survey response proportions.
(A) Correlations are shown for different language models: an N-gram language model finetuned on media, BERT without finetuning, and BERT finetuned on media.
The darker bars are computed using our \textit{synonym-grouping} method, which calculates media diet scores by grouping probabilities of synonyms of the target word. These results reinforce the importance of several of our modeling choices: leveraging a more powerful, pretrained language model (like BERT), synonym-grouping when computing target word probabilities, and adapting to media diet corpora.
Bootstrapped 95\% confidence intervals are shown.
(B) Regression analysis (error values and $R^2$ values) for predicting survey response proportions using the baseline BERT probabilities, the media diet scores, and the media diet score combined with the proportion of respondents paying very close attention to news. The media diet score achieves a statistically significant lower error and greater $R^2$. Attention to news is also an significant feature (see Table \ref{tab:pew_reg}).
(C) Correlations based on manually translated prompts (Orig) and automatically-generated paraphrases of those prompts (SynSub and BT). SynSub refers to synonym replacement -based paraphrasing, and BT refers to backtranslation-based paraphrasing. While there is some variation, the results are largely robust to specific paraphrases of the prompts used as inputs to language models.
(D) Correlations cut by specific media diets (FOX, CNN, NPR, NYT) and media sources (web news, TV, radio). 
For example, ``NPR-Radio'' is a BERT model adapted to NPR radio transcripts; there is a correlation of 0.49 between this model's scores and respondents who listed NPR as their primary news source.
Additional information on media characterization of each subpopulation is listed below the chart. For instance, 59\% of NYT respondents primarily obtain their news from the web. 
These results indicate that the approach is effective across sources and mediums. 
Moreover, the trend of greater correlation with more accurate media diet characterization, e.g. FOX-Web = 0.38 and 22\%, while NYT-Web = 0.54 and 59\%, suggests that more accurate media diet information for a subpopulation can lead to greater accuracy.
}
\normalsize
\label{fig:pew_main}
\end{figure}

We first find that the media diet models do have predictive power for public opinion prediction. We show correlations between model scores and survey response proportions, as well as regressions to predict survey response proportions.
The correlation between media diet scores and survey proportions is $r$=0.458, CI(0.350,0.553).
Regression results, shown in Table \ref{tab:pew_reg} and Figure \ref{fig:pew_main}D, indicate that the model score is a statistically significant feature ($\beta$=0.115, (0.087, 0.142)).

\begin{table}[!h]
\centering
\small
\caption{
Attitudes towards COVID-19: feature importances in regressions of media diet scores against survey response proportions. 
For one media diet subpopulation, the ``attention to news'' feature is the percentage of respondents who answered that they were paying ``very close'' attention to coronavirus-related news.
The demographic features are the proportion of respondents that fall within a given demographic bucket.
We list two models: (Model 1), which regresses solely the media diet score against the survey response proportions, and (Model 2) a version that includes attention to news and demographics covariates.
Statistically significant features are highlighted, and bootstrapped 95\% confidence intervals are shown in parentheses.
Media diet scores are significant for Model 1, and combination of these scores with the ``attention to news'' feature is both large in magnitude and significant in Model 2.
While there is undoubtedly a relationship between demographic buckets and media diets, demographic features alone cannot predict survey proportions and are not significant features in Model 2.
}
\normalsize
\small
\begin{tabular}{c c c}
\hline\hline 
Variable & \textbf{(Model 1) Media diet} & \textbf{(Model 2) Media diet + attention + demographics}  \\ 
\hline
$intercept$ & \cellcolor{blue!25} $0.194^{***}$ (0.134, 0.254) & 0.139 (-0.137, 0.416) \\
\textit{media diet score} & \cellcolor{blue!25}  $0.115^{***}$ (0.087, 0.142) & -0.204 (-0.428, 0.020) \\
\textit{attention to news} &  &  -0.192 (-1.154, 0.770) \\
\textit{(media diet score) $\cdot$ (attention to news)} &  & \cellcolor{blue!25} $0.523^{**}$ (0.164, 0.882) \\
$age1$ &  &  0.009 (-0.101, 0.118) \\
$age2$ &  &  0.119 (-0.055, 0.294) \\
$age3$ &  &  0.090 (-0.025, 0.206) \\
$age4$ &  &  -0.075 (-0.386, 0.235) \\
$edu1$ &  &  0.013 (-0.237, 0.264) \\
$edu2$ &  &  0.093 (-0.043, 0.230) \\
$edu3$ &  &  0.032 (-0.038, 0.102) \\
$race1$ &  &  0.017 (-0.239, 0.274) \\
$race2$ &  &  0.063 (-0.139, 0.265) \\
$race3$ &  &  0.056 (-0.118, 0.230) \\
$race4$ &  &  0.006 (-0.039, 0.051) \\
$sex1$ &  &  0.001 (-0.252, 0.255) \\
$sex2$ &  &  0.141 (-0.000, 0.283) \\
\hline  
R$^{2}$ &  0.2664  &  0.3327 \\
Error &  0.235 (0.225, 0.253) & 0.223 (0.215, 0.237) \\
\bottomrule
\multicolumn{3}{@{}l@{}}{\footnotesize Note: $^{*}\, p<0.05$; $^{**}\, p<0.01$; $^{***}\, p<0.001$}
\end{tabular}
\normalsize
\label{tab:pew_reg}
\end{table}

Our analyses support several of our modeling choices. First, we find that adapting the BERT model specifically is important. BERT by itself has a weak correlation with survey responses ($r$=0.274, CI(0.151, 0.388)). Though it hasn't been adapted to any news stories from 2020, its training data includes Wikipedia articles about previous epidemics, which imbues it with some knowledge about the threat of a coronavirus and possible precautionary steps. The ``Media Diet (N-gram)'' model, which adapts a non-neural language model to media diets, achieves a weaker correlation than the full BERT-based media model ($r$=0.237, CI(0.113, 0.354) vs. $r$=0.458, CI(03.50, 0.553)). This aligns with our thesis that language model-based media diet models can be linked to public opinion polls, though the language model must be sufficiently powerful to capture the semantics of media messages. 
Second, our synonym-grouping method (for computing probability-based media diet scores) helps produce significantly stronger correlations ($r$=0.458, CI(0.350,0.553) vs. $r$=0.190, CI(0.063,0.310)), shown in the darker bars in Figure \ref{fig:pew_main}A. Intuitively, this ``fuzzy'' grouping helps better capture the overall meaning of the target word. \cite{holtzman2021surface}
This also loosely parallels techniques in surveying that averages responses to related questions, rather than making conclusions based on single-item responses. \cite{berinsky2017measuring}

The predictive power of the media diets holds and is robust (1) even when demographic information of each subpopulation is included, (2) across mediums (online, TV, radio), and (3) to the specific phrasing of the prompts. 
Media diets are typically correlated with subpopulation demographics. Despite this, we find in our regression analyses that media diet scores are large and statistically significant compared to age, education, race, and gender information, as shown in Table \ref{tab:pew_reg}.
We also find similar effect sizes whether using models trained on online news, TV, or radio transcripts, as shown in Figure \ref{fig:pew_main}B. 
Finally, Figure \ref{fig:pew_main}C shows the results with the original prompts, compared against two paraphrase settings. In the first, prompts are paraphrased by synonym substitution. In the second, prompts are paraphrased using a  ``backtranslation'' approach by translating to a different language and then back to English; this  results in both word replacements and sentence structure changes. 
Overall, each setting produces similar moderate to strong correlations ($r$=0.500, CI(0.407,0.582) and $r$=0.406, CI(0.340,0.469) vs. $r$=0.458, CI(0.350, 0.553)). The backtranslation-based approach, which produces larger changes in the prompt, produces a slightly larger and negative difference, suggesting some sensitivity to major changes in phrasing.

To address RQ2 -- whether the model is sensitive to the amount of attention people are paying to news -- we include in our regression models the percentage of people who replied they were paying ``very close'' attention to coronavirus-related news. Table \ref{tab:pew_reg} shows that combining the model score with this attention value results in a particularly predictive, statistically significant feature ($\beta$=0.523, CI(0.164, 0.882)). The addition of this feature also increases the $R^{2}$ (0.3327 vs. 0.2664) and decreases the error (0.223, CI(0.214,0.237) vs. 0.235, CI(0.225, 0.253)).
This question is essentially a media \textit{exposure} question, and there is additional supporting evidence of the importance of this aspect.
For instance, 59\% of NYT respondents (i.e. people who selected NYT as their primary news source) stated that they get their news from the Web, compared to 22\% for FOX respondents. Thus, a model trained on web articles better represents NYT respondents. Correspondingly, the correlations are larger, with NYT-Web's $r$=0.541, CI(0.331,0.700) and FOX-Web's $r$=0.384, CI(0.142,0.583). There may or may not be significant differences, as the error bars overlap, but we note the trend holds for all outlets and mediums, suggesting that more accurate representations of one's media diet would lead to greater predictive accuracy.

\subsection*{Domain 2: Consumer Confidence}

\begin{figure}[!h]
\centering
\includegraphics[width=1.0\linewidth]{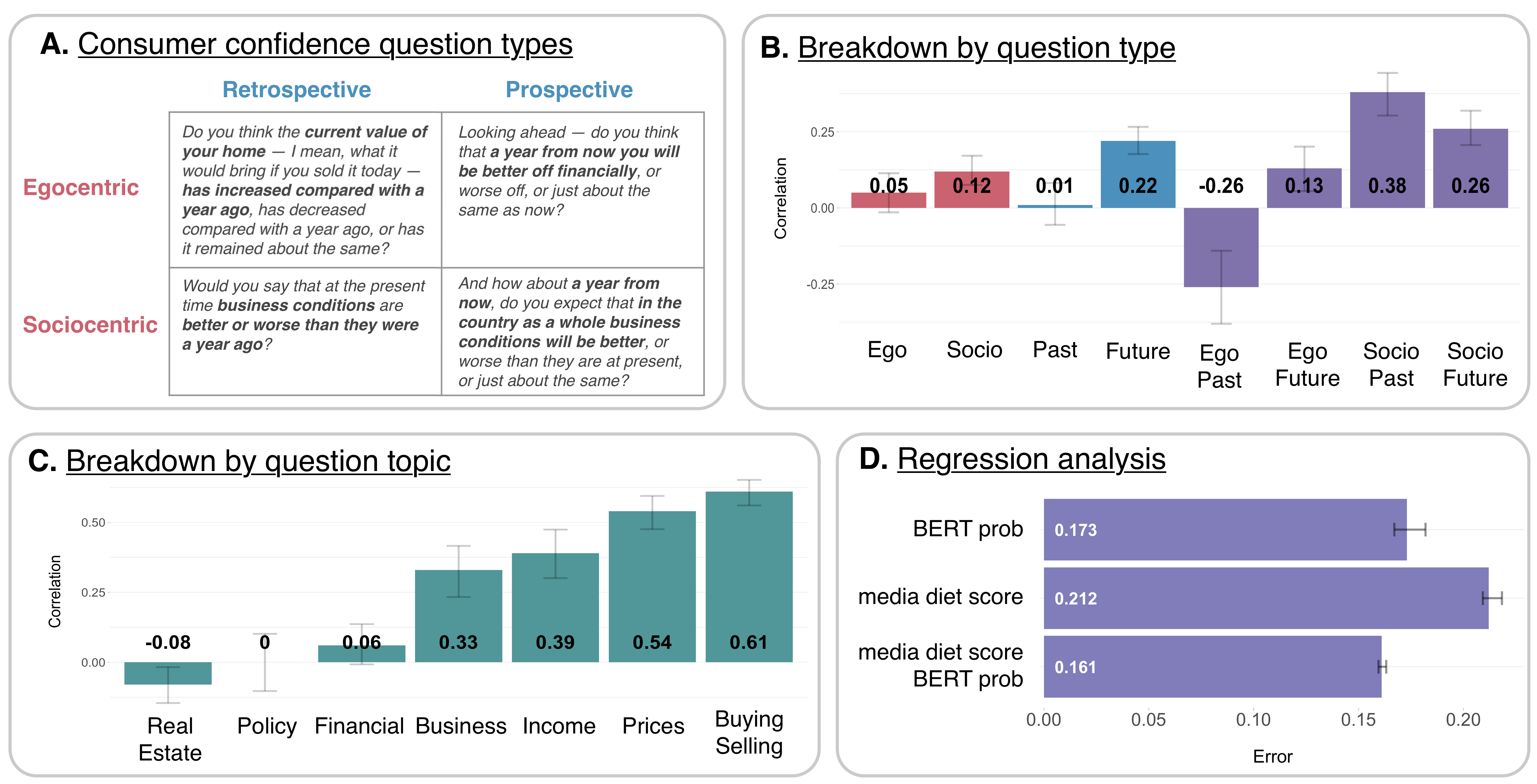}
\small
\caption{
Consumer confidence: correlations and regressions on media diet scores and survey response proportions.
(A) The survey questions are designed along two axes: egocentric (about one's personal situation) versus sociocentric questions (about the country or one's municipal), and retrospective (assessments relative to the past) versus prospective (assessments about the future) questions. An example prompt to a media diet model would be: `At the present time, business conditions are [BLANK] than they were a year ago.' with `better' and `worse' as target words.
(B,C) Correlations are split by question type or topic. Predictive signal is highly dependent on question type and topic, such as whether the survey question is polling about opinions about the future or the past. 
These results are aligned with literature on the types of opinions that are affected by media consumption.
(D) Regression against survey response proportions using the baseline BERT probabilities, the media diet scores, and the outputs of a general additive model combining the BERT probability and the media diet score.
A non-linear combination of media diet score and the original BERT probability improves against the BERT baseline. 
}
\normalsize
\label{fig:consumer_conf}
\end{figure}

The consumer confidence setting is helpful in answering RQ3 (which types of opinions are most affected by media consumption).
Across all questions, there is only a weak correlation ($r$=0.104, CI(0.066, 0.142)), but categorizing the questions by question type is illuminating.
The consumer confidence survey questions are designed to be either egocentric or sociocentric (personal vs. societal oriented), and retrospective or prospective (past reflection vs. future looking). An example of an egocentric, retrospective question is ``Compared with 5 years ago, do you think the chances that you will have a comfortable retirement have gone up, gone down, or remained the same?'' 
Figure 3A provides examples, and 
Figure \ref{fig:consumer_conf}B breaks out the results by question type. 
We see that the correlations are especially low for the retrospective questions ($r$=0.013, CI(-0.055, 0.082)), as well as the egocentric retrospective questions ($r$=-0.265, CI(-0.380, -0.141)). 
However, correlations for sociocentric-retrospective predictions are higher ($r$=0.376, CI(0.303, 0.443)), and sociocentric-prospective predictions ($r$=0.264, CI(0.206, 0.319)) are higher than egocentric-prospective predictions ($r$=0.0.129, CI(0.055, 0.201)).

- in the first paragraph we should also highlight that the socio-past category of predictions is strong (.38) and socio-future better than the ego-future.

Intuitively, answers to these kinds of questions are more likely to be a product of specific, personal situations, rather than being affected by news consumption.
These results also align with findings that show the tone of news coverage only affects sociocentric and prospective attributes. For instance, as people were exposed to more positive economic news, only their evaluations of \textit{the national economy} and expectations for the \textit{future} improved. \cite{boukes2019media}

The regression results are shown in Figure \ref{fig:consumer_conf}D. 
We find that the media diet scores do have predictive power, though we use a non-linear, general additive model to combine the scores with raw BERT scores ($error$=0.161, CI(0.160, 0.163) vs. $error$=0.173, CI(0.167, 0.182)). Similar to the COVID-19 setting, this is robust to paraphrasing.

The results here, considered jointly with the results in the COVID-19 setting, suggest several aspects of public opinion formation and media effects.
Earlier media effects research has emphasized the importance of considering the ``stability'' of prior beliefs, suggesting that political scientists will find larger media effects on issues or candidates for which people have weak prior opinions. \cite{bartels1993messages}
Figure \ref{fig:consumer_conf}C groups the questions into topic categories, and we see that effects are larger on more concrete topics, such as businesses, incomes, and goods. For broader, politically-dependent topics such as policy and financial issues, people may already have strong prior opinions. This notion of prior beliefs – and the novelty of the coronavirus – could also explain the stronger correlations found in the COVID-19 setting.
These results are also a reminder of the importance of accurate media diet characterization. Our media diet models are finetuned on news from the economy and finance sections, but people may be forming their opinions based on other sources. For example, `soft news' provided by late night entertainment talk show hosts, has been found to influence the amount of facts obtained on important issues such as foreign crises. \cite{baum2003soft} Research has also found that political and general news can also have an impact, with presidential dealings and ``extraordinary'' political events affecting consumer sentiment, even when controlling for economic conditions. \cite{10.2307/1519924}

\subsection*{Explaining predictions}

While deep neural networks perform a variety of tasks well, it can be difficult to interpret their behavior. 
A common question is: why did a model make a particular prediction? One approach to help answer this attempts to trace input-output behavior back to the original training examples. Methods that are faithful, or that accurately quantify the importance of individual training points, are typically too computationally expensive for large neural networks and moderately sized training or finetuning datasets. \cite{koh2017understanding} We thus utilize a simpler approach, wherein we find the nearest neighbors in BERT-embedding space between sentences in the media datasets used to adapt models, and a filled-in prompt. Nearest neighbors have previously been used to shrink the search space of more faithful methods, which suggest that they can be used to shed coarse insight into model behavior.
 Table \ref{tab:nn_minor_threat} shows the nearest neighbors for the completed prompt ``The coronavirus outbreak is a [minor] threat to the health of the U.S. population.'' 
Compared to the CNN model, the FOX model predicts more people would answer that it is a \textit{minor} threat, and this behavior is reflected in the original training set.
Inspecting these nearest neighbor results can help understand model behavior, as well as shed insight into the specific messages that may be important in opinion formation.

\begin{table}[!htb]
\caption{
Nearest neighbor analysis for understanding predictions of media diet models.
For the filled-in prompt ``The coronavirus outbreak is a [minor] threat to the health of the U.S. population.'', we show the top 10 most semantically similar sentences in the training sets for CNN/FOX media diet models.
The FOX model predicts a comparatively higher percentage of people would answer that it is a `minor' threat (as opposed to `major') relative to the CNN model.
Sentences with (\greencheck) imply agreement (roughly) with the completed prompt, i.e. they suggest that the outbreak is only a minor threat. Sentences with (\redxmark) contradict (roughly) this statement.
Correspondingly, a greater number of training examples in the FOX set are aligned with the statement.
}
\label{tab:nn_minor_threat}
\scriptsize
        \centering
        \begin{tabular}{p{8cm} | p{8cm} }
        \toprule
        \textbf{Nearest neighbors in CNN training set} & \textbf{Nearest neighbors in FOX training set} \\
        \midrule
        (\greencheck) The novel coronavirus outbreak is raging in China, but fewer than 1,000 people have been infected outside the country. 
        & (\greencheck) They have told us, 'most people are not at serious risk from coronavirus, and while anyone can get it and spread it so far, only a small number end up sick enough to be hospitalized.\\ \hline 
        
        (\greencheck) The bottom line, experts said, is that there is an extremely low risk of contracting coronavirus from the food supply. 
        & (\greencheck) For most people, the new coronavirus causes only mild or moderate symptoms.\\ \hline 
        
        (\redxmark) More than 8,500 people in the United States have been infected with coronavirus, and that number changes significantly by the hour.
        & (\redxmark) Look, the coronavirus is a serious health threat – no one can dispute that.\\ \hline 
        
        (\redxmark) There are more than 18,000 cases of coronavirus in the US.
        & (\greencheck) Coronavirus has raised a lot of stressful questions — how to cook, how to balance work and family, how to prevent the spread — but one seemingly positive note?\\ \hline 
        
        (\redxmark) Look we don't get very sick [from the coronavirus] but believe it or not we could still have that virus.
        & (\greencheck) Still, the new poll finds that about 3 in 10 Americans say they’re not worried at all about the coronavirus illness.\\ \hline 
        
        (\redxmark) The news comes amidst rising coronavirus numbers in the US.
        & (\greencheck) Americans will survive uncertainty of coronavirus – We are stronger than we realize.\\ \hline 
        
        (\greencheck) So as the coronavirus spreads , parents, teachers, caregivers and others have increasing concerns about how the disease affects them -- but there is some good news.
        & (\greencheck) Despite the more than 9,000 coronavirus cases in the U.S., the virus has largely spared children, which is puzzling as they are typically among the most vulnerable when it comes to seasonal illnesses like the flu or other coronaviruses.\\ \hline 
        
        (\redxmark) As we know now, the problem is that the specter of coronavirus is not only about you -- it's about everyone around you.
        & (\greencheck) The initial data from China showed that the effect of coronavirus on the heart is relatively small.\\ \hline 
        
        (\redxmark) Children's coronavirus cases are not as severe, but that doesn't make them less serious.
        & (\redxmark) The number of novel coronavirus cases in the U.S. has seen a 534 percent spike in just a week’s time as the amount of COVID-19 testing kits become more widely available across the country.\\ \hline 
        
        (\redxmark) From January until last week, Trump consistently minimized the risk the coronavirus posed to the country.
        & (\redxmark) The coronavirus pandemic has swept the globe, with more than 6,400 cases recorded in the United States to date.\\ \bottomrule
        \end{tabular}
\normalsize
\end{table}

\begin{figure}[!h]
\centering
\includegraphics[width=0.85\linewidth]{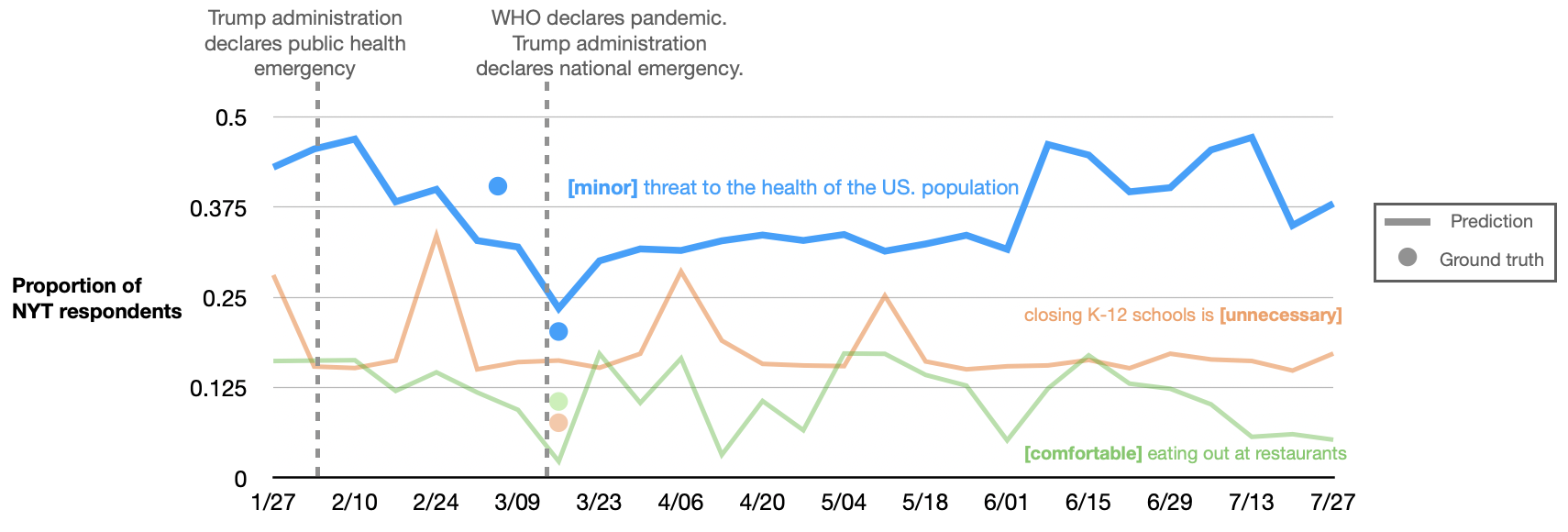}
\small
\caption{
Media diet models can be applied over time to supplement current surveys, forecast opinion, or retroactively measure sentiment. 
Predictions to three existing survey questions from weekly models trained on NYT datasets during 2020 are shown here. Ground truth proportions for surveys conducted during this time are shown as dots.
The model predicts that \textasciitilde45\% of NYT readers would say that ``the novel coronavirus is a \textit{minor} threat to the health of the U.S. population'' in late Jan 2020. This declines to \textasciitilde25\% over the first few months of the year, which reflects ground truth survey proportions and a shifting realization about the severity of the pandemic.
}
\normalsize
\label{fig:survey_over_time}
\end{figure}

\section*{Discussion}

These results illustrate that an AI system can predict human survey responses, by adapting a pretrained language model such as BERT to subpopulation-specific media diets. Earlier approaches using natural language processing to measure or forecast public opinion rely on much simpler summary statistics of media content. Our new approach leverages BERT’s ability to understand messaging at a deeper level, can function as a question-answering system, and can roughly connect answers back to specific media content. 
We demonstrate this in two separate, wide-reaching domains – public health and the economy – in which public opinion polls drive critical decision-making for public and private organizations alike.
If this method were refined and made more accurate,
this approach could supplement existing surveys, which are often cost-limited in the number of questions asked, the frequency conducted, and the subpopulations included.
The monthly survey-based Consumer Confidence index, for example, could be supplemented by a daily media diet model that predicts subpopulation response proportions to different questions. Figure \ref{fig:survey_over_time} illustrates one example of rolling, weekly models in the COVID-19 context.

Characterizing the fidelity of powerful language models is increasingly important, and there is an emerging field of study around empirically probing these models to understand their behavior.
For example, language models are now known to often reflect human biases in the training data, have issues with common sense reasoning, hallucinate false information, and face miscalibration issues. \cite{weidinger2021ethical}
Further research on explainability, such as methods that trace predictions back to training instances, or mechanistic interpretability of internal computations occurring within models, can help shed light on these issues.
These questions could also benefit from the social sciences. The science of polling, for example, faces several human-bias related measurement errors. These include authorship biases of questionnaire designers and social desirability biases of respondents. \cite{10.1086/517897,10.2307/2991824} Methodological advances in survey design \cite{10.2307/2998167, 10.2307/41403738} have helped address some of these, and there may be analogous approaches for ways to interact with language models in order to best elicit truthful responses.
More generally, these questions prompt the need for greater study of language modeling methods (e.g. for aligning with human intent), pitfalls of human-AI interaction (e.g. trust and automation bias), and societal implications (e.g. the potential for misuse of such models).

That these media diet models can predict survey responses at all raises several questions about the nature of public opinion \textit{formation}.
Early formulations focused on the apparent inconsistency of people's beliefs, which surfaced as response instability in people's answers to survey questions. Responses to the same questions, asked across time, often appeared as if randomly chosen. \cite{doi:10.1080/08913810608443650} Later models explained some of this instability by incorporating media effects. \cite{page1987moves, zaller1992nature, 10.2307/2111583} Zaller's ``Receive-Accept-Sample'' model, for example, posited that stated opinions are a function of received messages, how that information aligns with prior beliefs, and sampling from updated beliefs based on recency and saliency heuristics. \cite{zaller1992nature, tourangeau1988cognitive} The per-category consumer confidence results perhaps support this hypothesis about the importance of the ``stability’’ of prior beliefs. 
However, blind spots in public opinion formation models remain.
Research as early as Lazarsfeld in the 1950's, \cite{lazarsfeld1944people} as well as more recent work, \cite{valkenburg2013comm} has argued that estimating heterogeneous effects of media consumption requires requires systematic measures of how different people think.
This can include network effects, which are especially pertinent in the age of social media and have often not been incorporated.
Finally, we note that public opinion is only a window into privately held beliefs. 
Beliefs -- whether viewed from an epistemological lens as true/false statements, or as logical inferences from other beliefs, or as biochemical processes \cite{schwitzgebel2011belief} -- remains more difficult to study, though there has indeed been recent work in linking neural language models to human cognitive processes. \cite{gauthier2019linking}

Though public opinion prediction is our method's end goal, this approach also makes strides in the study of media effects. Meta-analyses of media effect size studies have typically found only small to moderate effect sizes, despite domain-expert expectations. For example, one analysis finds an average correlation of 0.16, and 90\% of correlations ranging from 0.07 to 0.32. A majority of these studies focus on behavioral change outcomes, such as physical activity or aggression, but small effect sizes are often found for knowledge and attitudinal changes as well. \cite{valkenburg2013comm, valkenburg2016media} Only for certain, specific outcomes such as agenda setting are media effects consistently moderate to large. \cite{luo2019meta} 
We argue the relatively large effect sizes found in the COVID-19 domain and certain consumer confidence categories is a natural consequence of our technique, which (a) leverages the representational capacity of BERT to model semantic content of media messages, and (b) connects those representations to specific questions through our probing method.
Several of our findings -- such as the persistence of predictive power when controlling for demographic features, or the significance of the `attention paid to news' feature in our regressions -- support the existence of media effects. 
However, we emphasize that our studies are not field experiments and do not measure casuality.

Three related problems reinforce the need for media diet -specific analysis: (1) selective exposure, or the general systemic bias in which people gravitate towards information that is congruent with their prior beliefs, \cite{10.1086/267513} (2) echo chambers, in which the selected environments amplify and strengthen opinions shared with like-minded individual, \cite{echochambers} and (3) filter bubbles, in which content curation and recommendation algorithms surface items based on users' past behaviors, again confirming the users' worldviews. \cite{pariser2011filter} 
Media diet models could help identify subpopulations being exposed to potentially harmful messaging. They also open up the potential for more sophisticated messaging effects studies, such as the differential effects of specific phrasing. While this has been studied in laboratory settings \cite{thibodeau2011metaphors} and in limited online settings, \cite{tan2014effect} the lack of tools has largely precluded media effects researchers from doing such an analysis. \cite{valkenburg2016media}

Finally, we emphasize that our results do not imply that (a) humans can be substituted by AI, or that (b) ground truth surveys and conversations with humans can be replaced with AI models.
Our work is in the tradition of new natural language processing tools that can summarize vast amounts of data and support decision-making done by humans.
We speculate that media diet models are one example of a potential wave of new tools that can power research in political science, social psychology, and computational social science. 
Ultimately, our goal is for these models to help address real-world problems in a human-centric fashion.

\section*{Methods}

\textbf{News datasets.} Coronavirus-related online news articles were aggregated, analyzed, and enriched by AYLIEN using AYLIEN’s News Intelligence Platform. \cite{aylien_dataset} The weekly primary outlet datasets (CNN, Fox, NYTimes, NPR) contain an average of 384.3 articles and 12563.6 sentences. The media bias groupings used in our divisive prompts analysis were obtained from Allsides Media Bias Ratings. \cite{allsides} 
TV show transcripts for CNN and Fox News, and radio transcripts for NPR, were obtained through Factiva. The TV/radio-based coronavirus models were adapted to datasets that averaged 62.4 transcripts and 12697.8 sentences. The consumer confidence models were adapted to datasets that avaeraegd 118.1 articles/transcripts and 20991.2 sentences.

\textbf{Language models.} We note that our choice of BERT was intentional, as it was trained on Wikipedia and the BooksCorpus only, and not any online data. There exist similar models such as RoBERTa that are generally more performant on NLP benchmarks, but these are often trained on an unknown mix of online web data; this can thus make it more difficult to interpret the results after adapting to online news.
BERT was adapted to each media diet dataset using a lightweight adaptation framework, \cite{pfeiffer2020adapterhub} which adapts BERT by inserting a limited number of trainable parameters. Compared to fine-tuning all of the original BERT parameters, this decreases training time, reduces the memory footprint, and lends itself to more flexible downstream combination of models (e.g. creating media diet models comprised of multiple single-outlet models), all while maintaining modeling accuracy. Each model was trained for 20 epochs using the Adam optimizer with an initial learning rate of $1e-4$, $\beta_1=0.9$, and $\beta_2=0.999$. Our non-neural baseline media diet models were  3-gram models with Kneser-Ney smoothing trained on the respective media diet datasets.

\textbf{Probing models.} 
The score for the final synonm-grouping method is calculated as:
$s = \frac{\sum_{w' \in w \bigcup synonyms(w) } MediaDietBERT(prompt, w')}{BERT(prompt, w)}$.
Synonyms were obtained through the Oxford Languages dictionary and manually checked to match the word sense and fit gramamatically.
To do the normalization for our N-gram media diet models, scores $s$ are divided by the probability of $w$ under the Google Web Trillion Word Corpus. 

\textbf{Prompt construction and paraphrases.} Each survey question was manually converted to a prompt by rewriting the question as a sentence, while minimizing any wording or structural changes. The majority of survey questions had two ``opposing'' choices (e.g. ``a major threat'' or ``a minor threat''), which we converted to target words (``major'', ``minor'') for our language models. For survey questions with more than two choices (``definitely true'', ``probably true'', ``probably not true'', ``not true''), we chose two opposing choices (e.g., ``probably', ``not''). Thus with two data points per question, each question is equally weighted. 

We test two methods for automatically generating paraphrases of our manually constructed prompts. The first method substitutes synonyms by replacing words with nearest neighbors in word embedding space. \cite{NIPS2013_9aa42b31} The second method utilizes ``back-translation'', which creates paraphrases by translating from English to a target language, and then back to English. This method results in both synonym-based replacements and structural changes. We use the English-Dutch and Dutch-English translation models provided by FairSeq. \cite{edunov2018understanding} Translation in the English-to-Dutch direction is done by generating 25 samples using top-k sampling with $k=20$. Back-transalation in the Dutch-to-English direction is done on each sample using greedy decoding.

\textbf{Survey datasets.} For the COVID-19 setting, each Pew Pathway surveys was conducted on a nationally representative sample of 12,648 respondents. 
In the correlation and regression analyses, there were a total of 32 questions, 2 answers per question, and 4 media groups for a total of N=256 data points. In the consumer confidence setting, the University of Michigan surveys are conducted on a nationally representative sample of at least 500 respondents each month. In the correlation and regression analyses, there were 528 questions (22 questions asked repeatedly over 24 weeks), 2 answers per question, and 4 media diet groups for a total of N=4224 data points.

The consumer confidence surveys do not contain any questions around respondents' media diet information. In order to perform our analysis, we create media diet groups by matching on demographics as following: (i) bucket Pew respondents according to four demographic factors (age, gender, region, education), (ii) compute which buckets have at least one ``primary'' news source, defined as any outlet for which at least 50\% of respondents use that outlet, (iii) combine buckets with the same set of news sources, (iv) compute consumer confidence responses per bucket.

\bibliography{main}

\end{document}